\pdfoutput=1
\documentclass[11pt]{article}
\usepackage[]{EMNLP2022}
\usepackage{times}
\usepackage{latexsym}
\usepackage[T1]{fontenc}
\usepackage[utf8]{inputenc}
\usepackage{microtype}
\usepackage{booktabs}
\usepackage{inconsolata}
\usepackage{graphicx}
\usepackage{amssymb}
\usepackage{CJKutf8}
\usepackage{amsmath}
\usepackage{amsthm}
\usepackage{multirow}
\usepackage{multicol}
\usepackage{tablefootnote}
\usepackage{hyperref}

\title{ReCo: Reliable Causal Chain Reasoning via Structural Causal Recurrent Neural Networks}

\author{Kai Xiong$^1$\footnotemark[1]\quad
Xiao Ding$^1$\footnotemark[2]\quad
Zhongyang Li$^2$\quad
Li Du$^1$\quad
Ting Liu$^1$ \\
{\bf Bing Qin}$^1$ \quad
{\bf Yi Zheng}$^2$ \quad
{\bf Baoxing Huai}$^2$\\
$^1$\normalsize{Research Center for Social Computing and Information Retrieval}\\[-.05cm]
\normalsize{Harbin Institute of Technology, China}\\[-.05cm]
$^2$\normalsize{Huawei Cloud, China}\\[-.05cm]
{\small\tt\{kxiong, xding, ldu, tliu, qinb\}@ir.hit.edu.cn}\\[-.05cm]
{\small\tt\{lizhongyang6, zhengyi29, huaibaoxing\}@huawei.com}}

\begin{document}
\begin{CJK}{UTF8}{gkai}
\maketitle

\renewcommand{\thefootnote}{\fnsymbol{footnote}}
\footnotetext[2]{Corresponding Author}
\footnotetext[1]{This work was conducted during the internship of Kai Xiong at Huawei Cloud}

\begin{abstract}
Causal chain reasoning (CCR) is an essential ability for many decision-making AI systems, which requires the model to build reliable causal chains by connecting causal pairs. However, CCR suffers from two main transitive problems: threshold effect and scene drift. In other words, the causal pairs to be spliced may have a conflicting threshold boundary or scenario.
To address these issues, we propose a novel \textbf{Re}liable \textbf{C}ausal chain reas\textbf{o}ning framework~(ReCo), which introduces exogenous variables to represent the threshold and scene factors of each causal pair within the causal chain, and estimates the threshold and scene contradictions across exogenous variables via structural causal recurrent neural networks~(SRNN). Experiments show that ReCo outperforms a series of strong baselines on both Chinese and English CCR datasets. Moreover, by injecting reliable causal chain knowledge distilled by ReCo, BERT can achieve better performances on four downstream causal-related tasks than BERT models enhanced by other kinds of knowledge.
\end{abstract}

\section{Introduction}
Causal chain reasoning aims at understanding the long-distance causal dependencies of events and building reliable causal chains. Here, \emph{reliable} means that events in the causal chain can naturally occur in the order of causal evolution within some circumstance based on the commonsense~\cite{roemmele2011choice}. Causal chain knowledge is of great importance for various artificial intelligence applications, such as question answering~\cite{asai2019learning}, and abductive reasoning~\cite{du2021learning}. Many studies focus on the reliability of causal pair knowledge but ignore that of causal chain knowledge, especially in the natural language processing~(NLP) community.

Previous works mainly acquire causal chain knowledge by first extracting precise causal pairs from text with rule-based~\cite{heindorf2020causenet,ijcai2020-guided} or neural-based~\cite{ding2019elg,10.1145/3366423.3380107} methods, then connecting these causal pairs into causal chains based on the textual or semantic similarity between events. However, this straightforward approach may bring some transitive problems~\cite{johnson2015causal}, leading to unreliable causal chains, which would hinder causal-enhanced models to get higher performances. For example, given a cause event: \emph{``playing basketball''}, and two candidate effect events: \emph{``gets a technical foul''} and \emph{``gets a red card''}, an unreliable causal chain~(\emph{``playing basketball''}$\rightarrow$\emph{``dispute''}$\rightarrow$\emph{``gets a red card''}) would mislead the model to choose the less plausible effect \emph{``gets a red card''}.
\begin{figure}[t]
    \centering
    \includegraphics[scale=0.505]{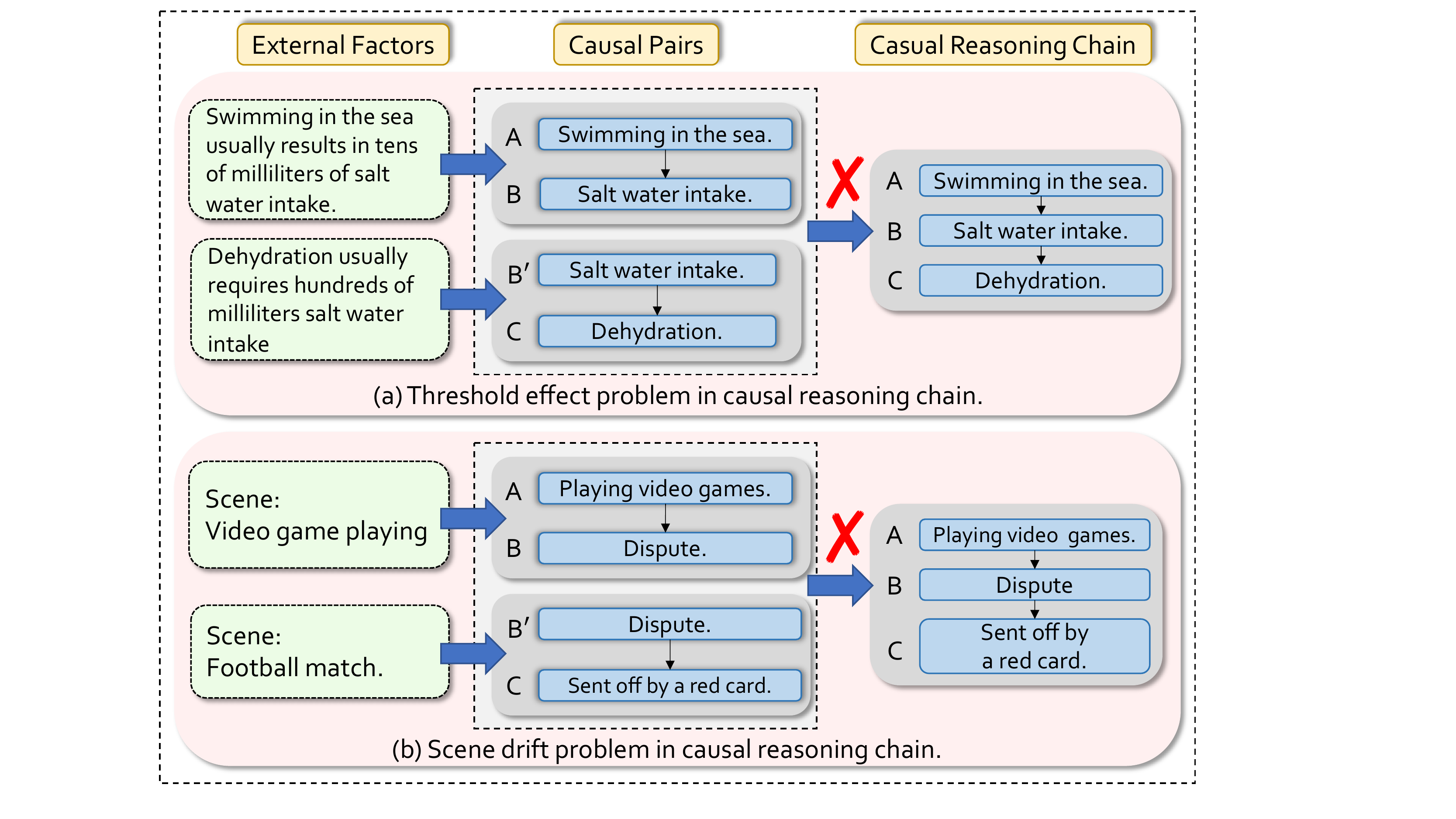}
    \caption{Causal chains with (a)~threshold effect and (b)~scene drift problems, which can be estimated by the contradictions of threshold and scene factors in the contexts, respectively.}
    \label{fig:instances}
\end{figure}

Among these transitive problems~\cite{johnson2015causal}, threshold effect and scene drift are the most two salient ones. As shown in Figure~\ref{fig:instances}~(a), given two causal pairs (A causes B, and B causes C), the threshold effect problem is that the influence of A on B is not enough for B to cause C. We can notice that, \emph{``swimming in the sea''} can only result in tens of milliliters of \emph{``salt water intake''}, while \emph{``dehydration''} is caused by hundreds of milliliters of \emph{``salt water intake''}. Therefore, \emph{``salt water intake''} conditioned on \emph{``swimming in the sea''} cannot lead to \emph{``dehydration''}. Similarly, as shown in Figure~\ref{fig:instances}~(b), the scene drift problem means that A $\rightarrow$ B and B $\rightarrow$ C would not happen 
within the same specific scene. These two \emph{``dispute''} events are wrongly joined together by their surface forms. \emph{``Dispute''} that happened in a video game scene cannot lead to \emph{``gets a red card''} in a football match scene. Therefore, we find that the threshold effect and scene drift problems are caused by the contradictions between the threshold factors and between the scene factors, respectively. 

To address these two issues, in ReCo, we first build a structural causal model~(SCM)~\cite{pearl2009causality} for each causal chain, and the SCM introduces exogenous variables to represent the threshold and scene factors of the causal pairs within the causal chain. Then, we conduct an exogenous-aware conditional variational autoencoder (EA-CVAE) to implicitly learn the semantic representations of exogenous variables according to the contexts of the causal pairs. Subsequently, we devise a novel causal recurrent neural network named SRNN to estimate the contradictions between the exogenous variables by modeling the semantic distance between them. Finally, we present a task-specific logic loss to better optimize ReCo.

Extensive experiments show that our method outperforms a series of baselines on both Chinese and English CCR datasets. The comparative experiments on different lengths of the causal chains further illustrate the superiority of our method. Moreover, BERT~\cite{devlin2019bert} injected with reliable casual chains distilled by ReCo, achieves better results on four downstream causal-related tasks, which indicates that ReCo could provide more effective and reliable causal knowledge. The code is available on \href{https://github.com/Waste-Wood/ReCo}{https://github.com/Waste-Wood/ReCo}.
\begin{figure}[t]
    \centering
    \includegraphics[scale=0.62]{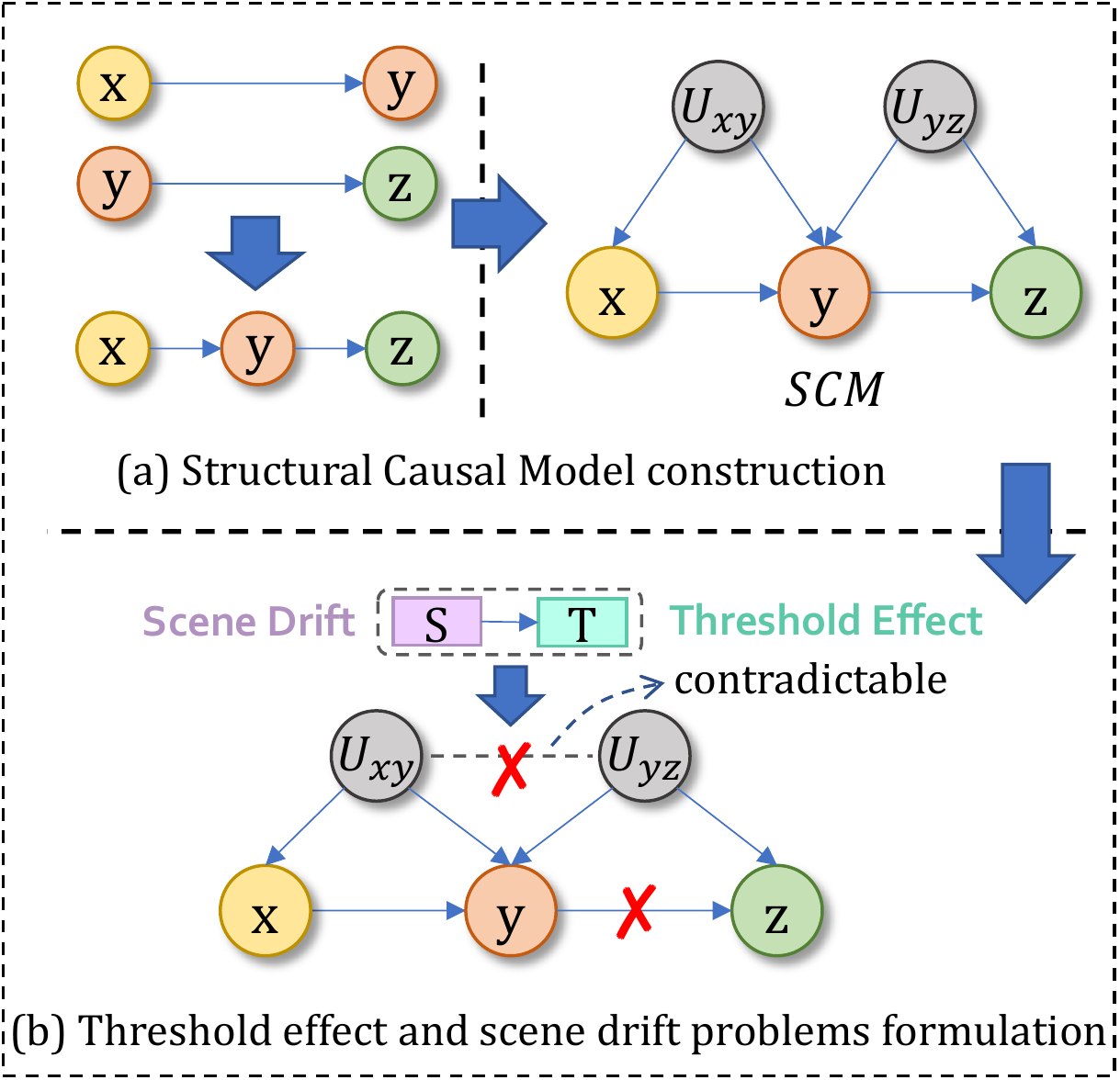}
    \caption{(a) Constructing SCM based on an antecedent causal chain and a causal pair. (b) If there is threshold effect or scene drift problem, then $U_{xy}$ would contradict $U_{yx}$. And it is worth discussing the threshold effect problem when scenes are consistent.}
    \label{fig:background}
\end{figure}
\vspace{-0.5cm}
\section{Background}

\subsection{Problem Definition}
In this paper, the CCR task is defined as a binary classification problem. Specifically, input a reliable antecedent causal chain ($x_1\rightarrow \cdots\rightarrow x_n$) and a causal pair ($x_n\rightarrow x_{n+1}$), the model needs to output whether the causal chain $x_1\rightarrow\cdots\rightarrow x_n\rightarrow x_{n+1}$ is reliable or not.

\subsection{Structural Causal Model}
Structural Causal Model~(SCM) was proposed by \citet{pearl2009causality}, which is a probabilistic graph model that represents causality within a single system. SCM is defined as an ordered triple $<U, V, E>$, where $U$ is a set of exogenous variables determined by external~(implicit) factors of the system. $V$ is a set of endogenous variables determined by internal~(explicit) factors of the system. $E$ is a set of structural equations, each structural equation represents the probability of an endogenous variable with the variables in $U$ and $V$.

\begin{figure*}[t]
    \centering
    \includegraphics[scale=0.51]{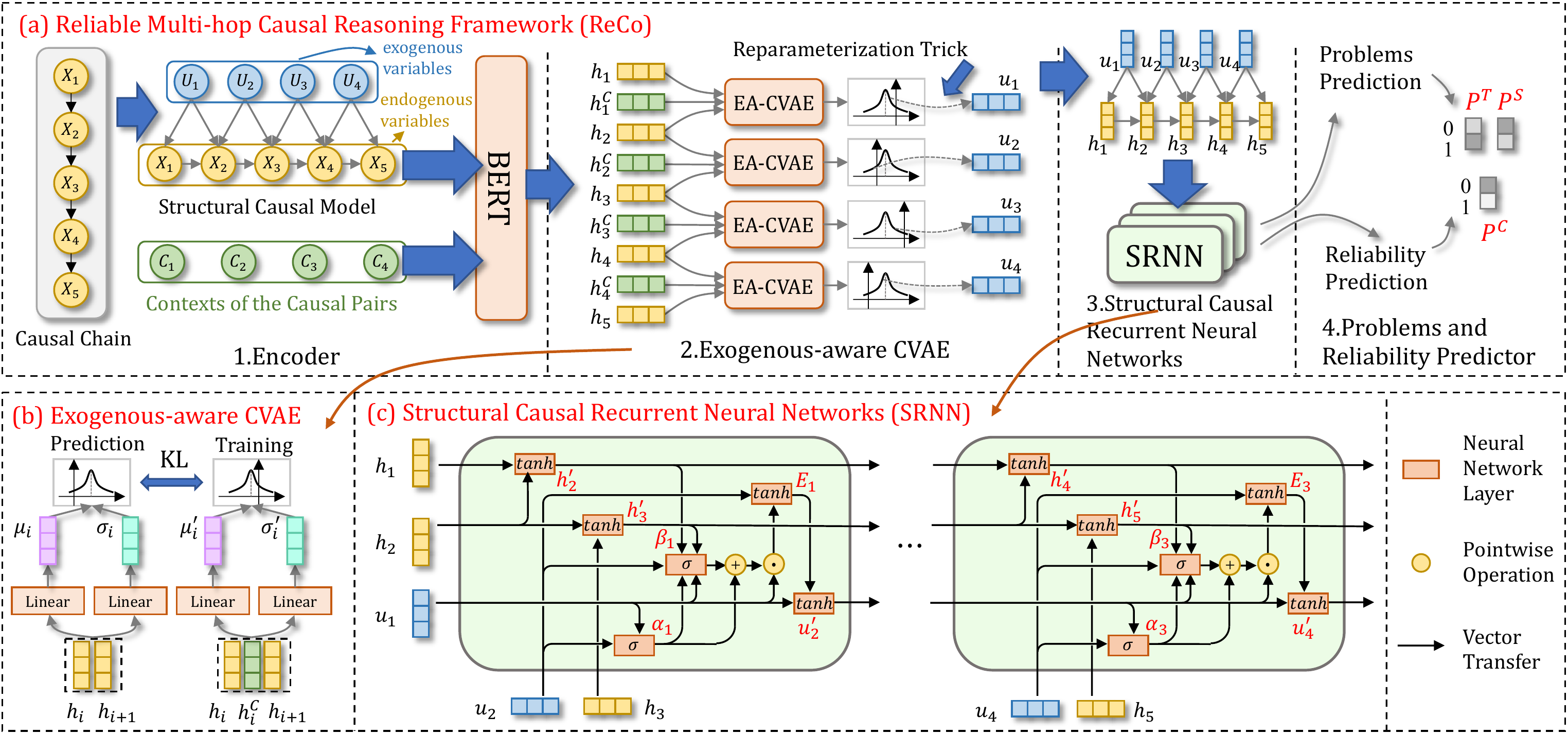}
    \caption{(a) The overall architecture of ReCo. (b) The detailed structure of EA-CVAE. (c) The detailed structure of SRNN which is a kind of recurrent neural networks.}
    \label{fig:framework}
    \vspace{-0.2cm}
\end{figure*}

As shown in Figure~\ref{fig:background}~(a), given two causal pairs~($x\rightarrow y$ and $y\rightarrow z$), they can be connected into a causal chain~($x\rightarrow y\rightarrow z$). We construct an SCM for this causal chain. Events~($x, y$ and $z$) are the endogenous variables, and exogenous variables~($U_{xy}$, $U_{yz}$) contain the threshold and scene factors of the causal pairs. Each structural equation represents the probability of an endogenous variable in $V=\{x, y, z\}$~(eg. $P$($y$|$x, U_{xy}$)). And as shown in Figure~\ref{fig:background}~(b), if the causal chain possesses threshold effect or scene drift problem, there are contradictions between $U_{xy}$ and $U_{yz}$.

\section{Method}

\subsection{Overview}
In this paper, we devise ReCo to estimate the reliability of the input causal chain. Figure~\ref{fig:framework}~(a) shows the architecture of ReCo, which consists of four components: (1)~an encoder to encode causal events and their contexts into dense vectors; (2)~an exogenous-aware CVAE to capture the exogenous variables with contexts; (3)~an SRNN to understand the causal chain along the direction of causality in the constructed SCM and solve the two transitive problems with two designed estimators; (4)~a predictor to predict the existence of the two transitive problems and the reliability of the causal chain. 

\subsection{Encoder}
Given a reliable antecedent causal chain ($X_1\rightarrow\cdots\rightarrow X_4$) and a causal pair ($X_4\rightarrow X_5$), the inputs of ReCo are a causal chain ($X_1\rightarrow\cdots\rightarrow X_5$) with 5 events and their 4 corresponding contexts ($C_1,\cdots, C_4$). $C_i$ denotes the context of the causal pair $X_i\rightarrow X_{i+1}$. We first construct an SCM for each causal chain, which introduces exogenous variables $U=\{U_1,\cdots, U_4\}$ to represent the threshold and scene factors of the causal pairs. The endogenous variables are the events $X=\{X_1, \cdots, X_5\}$ in the causal chain. Then we use BERT to encode the input events and contexts.

Specifically, we concatenate the events and their contexts into two sequences: \emph{[CLS] $X_1$ [SEP] $X_2$ [SEP] $X_3$ [SEP] $X_4$ [SEP] $X_5$ [SEP]}, and \emph{[CLS] $C_1$ [SEP] $C_2$ [SEP] $C_3$ [SEP] $C_4$ [SEP]}.
The final hidden states of \emph{[SEP]} tokens are set as the initial representations of the corresponding events and contexts. Then we scale them to $m$-dimension. Finally, we acquire event embeddings $H_X=\{h_1, h_2, h_3, h_4, h_5\}$ and context embeddings $H_C=\{h_1^C, h_2^C, h_3^C, h_4^C\}$, where $h_i, h_i^C\in\mathbb{R}^m$ denote the $i$-th event and context, respectively.

\subsection{Exogenous-aware CVAE}
Since the exogenous variables are hard to explicitly capture and CVAE has shown its ability to implicitly estimate variables~\cite{chen2021confounded,du2021learning}. Thus, we devise an EA-CVAE to capture the exogenous variables based on each causal pair and its corresponding contexts.

The EA-CVAE takes a causal pair and its corresponding context as inputs, and outputs the distribution of the exogenous variable. For example, as shown in Figure~\ref{fig:framework} (b), given a causal pair $h_i\rightarrow h_{i+1}$ and the corresponding context $h_i^C$, we first concatenate $h_i, h_{i+1}, h_i^C$ into $V_i=[h_i; h_{i+1}]\in \mathbb{R}^{2m}$ and $V^\prime_i=[h_i; h_i^C; h_{i+1}]\in \mathbb{R}^{3m}$. Hereafter, $V_i$ and $V^\prime_i$ are fed into the linear layers to estimate the mean and standard deviation values of the exogenous variable distribution:
\begin{equation}
\begin{aligned}
    &\mu_i=W_1V_i+b_1,\\
    &\sigma_i=\text{exp}(W_2V_i+b_2),\\
    &\mu^\prime=W_3V_i^\prime+b_3,\\
    &\sigma_i^\prime=\text{exp}(W_4V_i^\prime+b_4),
\end{aligned}
\end{equation}
where $W_1, W_2\in \mathbb{R}^{2m\times m}$ and $W_3, W_4\in \mathbb{R}^{3m\times m}$ are trainable parameters. The size of the multivariate normal distribution is set as $m$. Finally, we obtain two multivariate normal distributions $\mathcal{N}_i(\mu_i, \sigma_i^2)$ and $\mathcal{N}_i^\prime(\mu_i^\prime, \sigma^{\prime 2}_i)$.

After that, we conduct reparameterization trick to sample exogenous variables from $\mathcal{N}_i(\mu_i, \sigma_i^2)$ and $\mathcal{N}_i^\prime(\mu_i^\prime, \sigma^{\prime 2}_i)$. First, we sample a value $\epsilon$ from the standard normal distribution $\mathcal{N}(\mathbf{0}, \mathbf{I})$, and then we obtain the representation $u_i\in\mathbb{R}^m$ of the exogenous variable $U_i$ based on $\epsilon$:
\begin{align}
    u_i=
    \begin{cases}
        \mu_i^\prime + \epsilon\times \sigma^\prime_i  \quad \text{training},\\
        \mu_i+\epsilon\times \sigma_i \quad \text{prediction}.
    \end{cases}
\end{align}

Hereafter, for each causal pair, we get the representation of its corresponding exogenous variable and obtain $u=\{u_1, u_2, u_3, u_4|u_i\in \mathbb{R}^m\}$.
In the training stage, $u_i\in u$ is sampled from $\mathcal{N}_i^\prime(\mu_i^\prime, \sigma^{\prime 2}_i)$. While in the prediction stage, $u_i$ is sampled from $\mathcal{N}_i(\mu_i, \sigma_i^2)$. Thus, contexts are not required as inputs in the prediction stage.

Finally, we obtain the representations of the endogenous and exogenous variables in the SCM.




\subsection{Structural Causal Recurrent Neural Networks}

We propose SRNN to measure the reliability of the causal chain, and estimate the two transitive problems by measuring the semantic distance between the exogenous variables. As shown in Figure~\ref{fig:framework} (c), the SRNN consists of the following five components. The input of the SRNN in the first recurrent step is a quintuple $<h_1, h_2, h_3, u_1, u_2>$.

\paragraph{Scene Drift Estimator} We design this component to estimate the scene drift problem between two exogenous variables:
\begin{align}
    \alpha_1=\sigma(W_{m1}u_1+b_{m1}-W_{m2}u_2-b_{m2}),
\end{align}
where $\alpha_1\in \mathbb{R}^m$ is the measurement of the scene drift problem. $W_{m1}, W_{m2}\in \mathbb{R}^{m\times m}$ are trainable parameters, and $\sigma$ is the sigmoid function.

\paragraph{Hidden Gate} Hidden gate is used for aggregating the information within the endogenous variables for the next recurrent step of the SRNN and estimating the threshold effect problem:
\begin{align}
\label{eq:hidden}
    h_2^\prime = \tanh{(W_h[h_1; h_2]+b_h)},\\
    h_3^\prime = \tanh{(W_h[h_2; h_3]+b_h)},
\end{align}
where $h_2^\prime, h_3^\prime\in\mathbb{R}^{m}$ are the aggregated endogenous variables, and $W_h\in \mathbb{R}^{2m\times m}$ is a trainable parameter.

\paragraph{Threshold Effect Estimator} Since the threshold effect problem can be discussed iff the scene is consistent, and threshold factors are event-specific, we can estimate the threshold effect problem with the endogenous and exogenous variables based on the result of the scene drift estimator.
\begin{align}
\label{eq:beta}
    \beta_1\!=\!\sigma(W_{\beta}([u_2;h_3^\prime]\!-\![u_1;h_2^\prime])\!\odot\!(1-\alpha_1)),
\end{align}
where $\beta_1\in\mathbb{R}^m$ estimates whether the threshold effect problem exists, and $W_{\beta}\in\mathbb{R}^{2m\times m}$ is a trainable parameter.

\paragraph{Exogenous Gate} $u_1$ contradicts $u_2$ if there is threshold effect or scene drift problem. We can learn the contradiction of $u_1$ on $u_2$ by:
\begin{equation}
\label{eq:e7}
    E_1=\tanh(W_{e}(u_2+\frac{\alpha_1+\beta_1}{2}\odot u_1)+b_e),
\end{equation}
where $E_1\in\mathbb{R}^m$ is the representation of the contradiction of $u_1$ on $u_2$, and $W_{e}\in \mathbb{R}^{m\times m}$ is a trainable parameter. 
If there are not threshold effect and scene drift problems, $\alpha_1$ and $\beta_1$ are equal to 0, and $E_1$ is close to $u_2$.

\paragraph{Output Gate} For the inputs of the next recurrent step of the SRNN, we compose $u_1$ and $E_1$ into $u_2^\prime$.
\begin{equation}
\label{eq:out}
\begin{aligned}
    &u_2^\prime=\tanh(W_{o}[u_1;E_1]+b_{o}),\\
    \end{aligned}
\end{equation}
where $u_2^\prime\in\mathbb{R}^{m}$ is the aggregated exogenous variable, and $W_{o}\in\mathbb{R}^{m\times m}$ is a trainable parameter.

Finally, we denote $<h_2^\prime, h_3^\prime, h_4, u_2^\prime,$ $u_3>$ as the input to the next recurrent step of the SRNN.

\subsection{Problems and Reliability Predictor}
After the SRNN, we can obtain the final output $<\alpha_3, \beta_3, h_4^\prime, h_5^\prime, E_3>$. First, we can measure the existence of the threshold effect and scene drift problems based on $\beta_3$ and $\alpha_3$, respectively:
\begin{equation}
\begin{aligned}
    P^T&=\text{Softmax}(W_T\beta_3+b_T),\\
    P^S&=\text{Softmax}(W_S\alpha_3+b_S),
\end{aligned}
\end{equation}
where $P^T=[P_0^T; P_1^T],P^S=[P_0^S; P_1^S]\in\mathbb{R}^2$ are the probability distributions of threshold effect and scene drift problems, respectively. The subscript $0$ and $1$ of $P^T$ and $P^S$ denote the non-existence and existence probabilities of the corresponding problems.
$W_S, W_T\in\mathbb{R}^{m\times 2}$ are trainable parameters. Therefore, we can explain why this causal chain breaks according to $P^T$ and $P^S$.

Finally, we can measure the reliability of the causal chain as follows:
\begin{equation}
\begin{aligned}
    P^1&=\tanh(W_1[h_4^\prime; u_3]+b_1),\\
    P^2&=\tanh(W_2[h_5^\prime; E_3]+b_2),\\
    P^C&=\text{Softmax}(W_C[P^1;P^2]+b_C),
\end{aligned}
\end{equation}
where $P^1, P^2\in\mathbb{R}^m$ are the intermediate parameters, $P^C=[P_0^C; P_1^C]\in\mathbb{R}^2$ is the probability distribution of the reliability of the causal chain $X_1\rightarrow \cdots\rightarrow X_5$, and $P_0^C, P_1^C\in\mathbb{R}^1$ denote the probabilities that the causal chain is unreliable and reliable, respectively. $W_1, W_2\in\mathbb{R}^{2m\times m}$ and $W_C\in\mathbb{R}^{m\times 2}$ are trainable parameters. 

\subsection{Optimizing with a Logic Loss}

We design a logic loss to reduce the loss function from 4 parts to 3 parts. For example, if the causal chain is reliable, the probabilities that the two problems do not exist and the causal chain is reliable should be equal.
Therefore, the logic loss is:
\begin{equation}
L_{\text{Logic}}= |\log(P^T_0 \times P^S_0)-\log(P^C_1)|,
\end{equation}
where  $P_o^T$ and $P_0^S$ are the probabilities that the threshold effect and scene drift problems do not exist, and $P^C_1$ is the probability that the causal chain is reliable. Moreover, if the causal chain is unreliable due to the scene drift problem, the logic loss is $L_{\text{Logic}}=|\log(P^S_1\times P^T_0)-\log(P^C_0)|$.

Finally, the loss function is denoted as:
\begin{equation}
\label{eq:11}
\begin{aligned}
    &L = L_{\text{Chain}} + \lambda_1 L_{\text{Logic}} + \lambda_2 L_{\text{kl}},\\
    &L_{\text{Chain}}=\text{CrossEntropy}(Y, P^C),\\
    &L_{\text{kl}}= \sum_{i=1}^{4} \text{\text{KL}}(\mathcal{N}(\mu_i, \sigma_i^2) || \mathcal{N}(\mu_i^\prime, \sigma_i^{\prime2})),
\end{aligned}
\end{equation}
where $L_{\text{Chain}}$ is the loss of the causal chain reliability. $L_{\text{Logic}}$ is the logic loss. $L_{\text{kl}}$ is the Kullback-Leibler divergence loss~\cite{hershey2007approximating} of the EA-CVAE. $\lambda_1$ and $\lambda_2$ are loss coefficients.

\section{Experiments}
\subsection{CCR Datasets Construction}
We choose the Chinese causal event graph CEG~\cite{ding2019elg} and English CauseNet~\cite{heindorf2020causenet} to obtain unlabeled causal chain reasoning examples. 

We first use Breadth-First Search on CEG and CauseNet to retrieve 2,911 and 1,400 causal chains with contexts, respectively. Each causal chain has 5 events, and no more than three events are overlapped between any two causal chains. 

Then, we label the causal chains through crowd-sourcing. Professional annotators are asked to label the first causal relationship where the causal chain breaks, and which problem (threshold effect or scene drift) causes this break.
Each chain will be labeled by three annotators, the Cohen's agreement scores are $\kappa$ = 78.21\% and 75.69\% for Chinese and English CCR datasets, respectively.

We split the causal chains into different lengths of training examples (Instance-3, Instance-4, Instance-5). If a causal chain of length 5 breaks at the third causal relationship, 1 positive and 1 negative training examples are constructed (positive: $X_1\rightarrow X_2\rightarrow X_3$; negative: $X_1\rightarrow X_2\rightarrow X_3\rightarrow X_4$). The statistics of the two CCR datasets are shown in Table~\ref{tab:data}. Refer to Appendix~\ref{sec:examples} for Chinese and English CCR examples.

\begin{table}[t]
\small
\centering
\begin{tabular}{llrrr}
\toprule
\multicolumn{2}{c}{\textbf{CCR}}                          & \textbf{Train} & \textbf{Dev} & \textbf{Test}  \\ \midrule
\multicolumn{1}{l|}{\multirow{6}{*}{\textbf{Zh}}} 
& Chain     & 2,131  & 290  & 490  \\
\multicolumn{1}{l|}{}  & Instance-3 & 2,131  & 290  & 490  \\
\multicolumn{1}{l|}{}  & Instance-4 & 1,552  & 207  & 324  \\
\multicolumn{1}{l|}{}  & Instance-5 & 1,077  & 139  & 188  \\
\multicolumn{1}{l|}{}  & Total & 4,760 & 636 & 1,002 \\ \midrule
\multicolumn{1}{l|}{\multirow{6}{*}{\textbf{En}}} 
& Chain     & 1,037  & 139  & 224  \\
\multicolumn{1}{l|}{}  & Instance-3 & 1,037  & 139  & 224  \\
\multicolumn{1}{l|}{}  & Instance-4 & 829   & 109  & 164  \\
\multicolumn{1}{l|}{}  & Instance-5 & 612   & 80  & 105   \\
\multicolumn{1}{l|}{}  & Total      & 2,478  & 328  & 493  \\ \bottomrule
\end{tabular}
\caption{Statistics of CCR datasets. Chain denotes the causal chains retrieved from causal event graphs.  Instance-3, Instance-4 and Instance-5 denote the instance with chain lengths of 3, 4 and 5, respectively.}
\label{tab:data}
\end{table}

\subsection{Baselines}
We compare the performance of ReCo against a variety of sequence modeling methods, and causal reasoning methods developed in recent years. In Embedding and ExCAR, for a causal chain $X_1\rightarrow\cdots X_n$, we treat $X_1\rightarrow\cdots X_{n-1}$ and $X_n$ as the cause and effect, respectively.
\paragraph{Embedding}\cite{xie2019distributed} measures word-level causality through causal embedding. We choose the max causality score between cause and effect words, and apply a threshold for prediction.

\paragraph{LSTM}\cite{hochreiter1997long} is a recurrent neural network. We use BiLSTM to represent the causal chains for binary classification.

\paragraph{BERT}\cite{devlin2019bert,cui-etal-2020-revisiting} is pre-trained unsupervised with massive unlabeled data. Specifically, we use BERT-base to represent the causal chains for the reliability classification.

\paragraph{ExCAR}\citep{du2021excar} introduces evidence events for explainable causal reasoning. We introduce evidence events to the cause-effect pair for ExCAR experiments.

\paragraph{CausalBERT}\citep{li2021causalbert} injects massive causal pair knowledge into BERT. CausalBERT is used to represent the causal chain for experiments.

We use precision, recall, F1 score, and accuracy to measure the performance of each method.

\subsection{Training Details}
For ReCo, we use the pre-trained BERT-base~\cite{devlin2019bert,cui-etal-2020-revisiting} as the encoder to encode events and contexts. The batch size is set to 24, the dimension $m$ is 256, we choose Adam~\cite{kingma2014adam} as the optimizer with a learning rate of 1$e$-5. The loss coefficients $\lambda_1$ and $\lambda_2$ are 1 and 0.01, respectively. ReCo runs 50 epochs on two Tesla-P100-16gb GPUs.

\begin{table}[t]
\small
\centering
\setlength\tabcolsep{5.3pt}
\begin{tabular}{llcccc}
\toprule
\textbf{CCR}                                           & \textbf{Methods} & \textbf{P}     & \textbf{R}     & \textbf{F1}    & \textbf{Acc \%} \\ \midrule
\multicolumn{1}{l|}{\multirow{6}{*}{\textbf{Zh}}} 
& Embedding & 61.30 & 82.75 & 70.43 & 58.18             \\
\multicolumn{1}{l|}{}                                  
& LSTM & 63.64 & 83.58 & 72.26 & 61.38             \\
\multicolumn{1}{l|}{}                                  
& BERT & 64.85 & 86.90 & 74.27 & 63.77             \\
\multicolumn{1}{l|}{}                                  
& ExCAR & 63.97 & 86.57 & 73.57 & 62.57             \\
\multicolumn{1}{l|}{}                                  
& CausalBERT & 64.53 & 87.23 & 74.19 & 63.47             \\
\multicolumn{1}{l|}{}                                  
& ReCo (Ours) & \textbf{66.50} & \textbf{87.56} & \textbf{75.59} & \textbf{65.97}    \\ \midrule
\multicolumn{1}{l|}{\multirow{6}{*}{\textbf{En}}} 
& Embedding & 65.30 & 81.17 & 72.55 & 59.63             \\
\multicolumn{1}{l|}{}                                  
& LSTM & 71.13 & 85.19 & 77.53 & 67.55             \\
\multicolumn{1}{l|}{}                                  
& BERT & 72.75 & 84.88 & 78.35 & 69.17             \\
\multicolumn{1}{l|}{}                                  
& ExCAR & 73.33  & 84.88 & 78.68 & 69.78           \\
\multicolumn{1}{l|}{}                                  
& CausalBERT & 72.38 & 87.35 & 79.16 & 69.78             \\
\multicolumn{1}{l|}{}                                  
& ReCo (Ours) & \textbf{74.03} & \textbf{87.96} & \textbf{80.39} & \textbf{71.81}    \\ \bottomrule
\end{tabular}
\caption{Overall results on the CCR test sets.}
\label{tab:results}
\end{table}

\subsection{Overall Results}
We implement Embedding, LSTM, BERT, ExCAR, CausalBERT and ReCo on both Chinese and English CCR datasets. The overall results are shown in Table~\ref{tab:results}, from which we can observe that:

(1) Comparing word-level method (Embedding) to event-level methods (LSTM, BERT, ExCAR, CausalBERT and ReCo), event-level methods achieve absolute advantages, which indicates that considering the causality between words and ignoring the semantics of events is not better for CCR.

(2) Knowledge-enhanced methods~(ExCAR and CausalBERT) achieve comparable results to BERT. This is mainly because not all the evidence events in ExCAR are reliable, and CausalBERT only possesses causal pair knowledge, making ExCAR and CausalBERT struggle with the CCR tasks.

(3) ReCo outperforms BERT, ExCAR and CausalBERT in F1 score and accuracy, which shows that the exogenous variables captured by the EA-CVAE are significant to conducting CCR tasks and the SRNN is important to address the two transitive problems. Moreover, the advantage of ReCo is mainly reflected in precision, it is because capturing the threshold and scene factors is effective to measure the transitive problems and estimate the reliability of the causal chains.

(4) All six methods get lower precision scores on the Chinese CCR test set than that on the English CCR test set. This is mainly because all events in the Chinese CCR dataset are sentences, and most of the events in the English CCR dataset consist of only one word, making the Chinese CCR task more challenging and more complex.

\begin{figure}[t]
    \centering
    \includegraphics[scale=0.47]{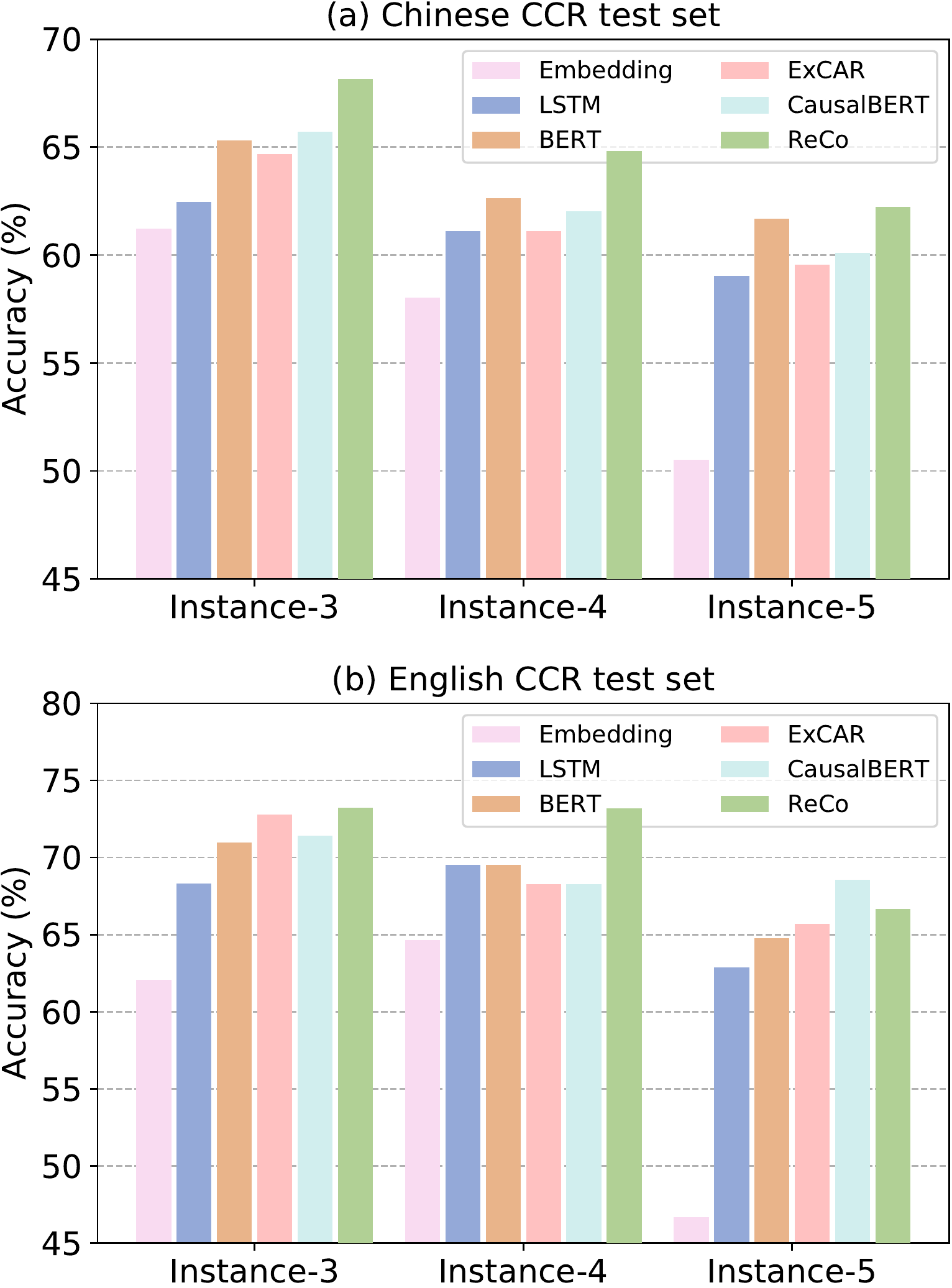}
    \caption{Accuracy on (a) Chinese and (b) English CCR test sets categorized by the lengths of the causal chains.}
    \label{fig:cnen}
\end{figure}

Moreover, we also compare ReCo with baselines on different lengths of causal chains. Results are illustrated in Figure~\ref{fig:cnen}. We can observe that:

(1) Most of the methods perform worse as the chain gets longer. It indicates that longer instances need stronger CCR ability. 

\begin{table*}[t]
\small
\centering
\begin{tabular}{lcccc}
\toprule
\textbf{Datasets}                         & \textbf{BERT$_\text{O}$}  & \textbf{BERT$_\text{P}$}  & \textbf{BERT$_\text{C}$}  & \textbf{BERT$_\text{R}$}  \\ \midrule
Event StoryLine v0.9\tablefootnote{Only the intra-sentence event pairs are kept for experiments and the train, dev, test sets are split randomly. We also ensure the cause event precedes the effect event.} \cite{caselli2017event} (\textbf{F1 \%}) & 66.84 & 68.08 & 69.05 & \textbf{70.66} \\
BeCAUSE~2.1 \cite{dunietz2017because} (\textbf{Accuracy \%})     & 79.17 & 81.94 & 83.33 & \textbf{83.80} \\
COPA \cite{roemmele2011choice} (\textbf{Accuracy \%})                & 73.80 & 74.00 & 74.20 & \textbf{75.40} \\
CommonsenseQA \cite{talmor2019commonsenseqa} (\textbf{Accuracy \%})       & 54.71 & 54.87 & 55.04 & \textbf{55.12} \\ \bottomrule
\end{tabular}
\caption{Overall results of causal knowledge injection. The evaluation metrics are computed based on manually split test sets (Event StoryLine~v0.9, BeCAUSE 2.1), official test (COPA) and dev (CommonsenseQA) sets.}
\label{tab:contrast}
\end{table*}

(2) ReCo performs best on almost all instance levels of both CCR datasets. This is mainly because the threshold and scene factors captured by the EA-CVAE are important for CCR tasks, and the SRNN can properly capture the two transitive problems by estimating the semantic distance between threshold factors or scene factors. However, CausalBERT achieves the best performance on the instance-5 of the English CCR test set. This is mainly because Instance-5 in the English CCR dataset might rely more on massive external causal pair knowledge.

(3) Compared with the results on Instance-4, results drop more on the English Instance-5 than that on the Chinese Instance-5. The reason is that conducting CCR on the causal chains with five or more word-level events might need more extra information to reason from the first event to the last event.

\subsection{Causal Knowledge Injection}

\begin{table}[t]
\small
\centering
\begin{tabular}{lc}
\toprule
\textbf{Methods} & \textbf{Accuracy \%} \\ \midrule
\textbf{BERT}  & 69.17    \\ 
-w context     & 69.37   \\\midrule
\textbf{ReCo}  & \textbf{71.81}    \\
-w/o EA-CVAE   & 68.97    \\
-w/o Problems Estimators & 70.18    \\
-w/o Logic Loss &70.18 \\\bottomrule
\end{tabular}
\caption{Overall results of the ablation study on the English CCR test set. ``w'' and ``w/o'' denote ``with'' and ``without'', respectively.}
\label{tab:alb2}
\end{table}

To further investigate the effectiveness of ReCo, we inject different kinds of causal knowledge into BERT. Then following \citet{du-etal-2022-e}, we test the causal-enhanced BERT models on four NLP benchmark datasets: a causal extraction dataset Event StoryLine~v0.9~\cite{caselli2017event}, two causal reasoning datasets BeCAUSE~2.1~\cite{dunietz2017because} and COPA~\cite{roemmele2011choice}, as well as a commonsense reasoning dataset CommonsenseQA~\cite{talmor2019commonsenseqa}. To give a careful analysis, we inject causal knowledge into BERT in the following four different ways (The details of knowledge injection can refer to Appendix~\ref{app:cki}):

$\bullet$ BERT$_{\text{O}}$ injected with no external knowledge.

$\bullet$ BERT$_{\text{P}}$ injected with causal pair knowledge.

$\bullet$ BERT$_{\text{C}}$ injected with unfiltered causal chain knowledge.

$\bullet$ BERT$_\text{R}$ injected with causal chain knowledge distilled by ReCo.

The results are shown in Table~\ref{tab:contrast}, from which we can observe that:

(1) Methods~(BERT$_\text{P}$, BERT$_\text{C}$, BERT$_\text{R}$) enhanced with causal knowledge outperform the original BERT~(BERT$_\text{O}$) on all four tasks, which indicates that causal knowledge can provide extra information to conduct causal-related tasks.

(2) Comparisons between causal chain knowledge enhanced methods (BERT$_\text{C}$, BERT$_\text{R}$) and causal pair knowledge enhanced method (BERT$_\text{P}$) show that BERT$_\text{C}$ and BERT$_\text{R}$ can push the model to a higher level than BERT$_\text{P}$ on all four tasks. The main reason is that causal chains contain more abundant knowledge than causal pairs.

(3) Unfiltered causal chain knowledge enhanced method (BERT$_\text{C}$) performs worse than BERT$_\text{R}$ injected with causal chain knowledge distilled by ReCo. The main reason is that some unfiltered causal chains would be unreliable due to the threshold effect or scene drift problem, which would mislead the model to choose the wrong answer.

\subsection{Ablation Study}
We provide ablation studies to show the superiority and effectiveness of ReCo. First, we provide the contexts of the causal pairs to BERT to prove the advantages of the EA-CVAE and SRNN in ReCo. Second, we remove the EA-CVAE in ReCo and set the contexts as the exogenous variables to investigate the effect of the EA-CVAE. Third, we remove the extra supervised signals of the problem estimators to study the effect of the problem estimators. Finally, we replace the logic loss with cross-entropy losses to validate the effectiveness of the logic loss. Overall results are shown in Table~\ref{tab:alb2}. From which we can find that:

(1) After providing contexts to BERT, the performance of BERT increases slightly, which shows that there is effective information in the contexts to conduct CCR tasks, but BERT cannot use it sufficiently. This proves that properly utilizing information in the contexts is of great importance.

(2) After removing EA-CVAE, the performance of ReCo drops and is worse than BERT. This is because the contexts are not proper estimations of the exogenous variables and there is also noise in the contexts which has negative impacts on ReCo. 

(3) After removing the supervised signals of the problem estimators, RoCo performs worse, it indicates the problem estimators supervised by the extra problem labels are important to measure the existence of the two transitive problems. Moreover, ReCo without the extra supervised signals outperforms contexts-enhanced BERT, which indicates the EA-CVAE in ReCo can properly estimate the exogenous variables with contexts, and the SRNN in ReCo plays an important role in deeply understanding the causal chain.

(4) After replacing the logic loss with cross-entropy losses, the performance of ReCo drops 1.63 in accuracy, which indicates that the logic constraints applied by the logic loss can guide ReCo to better generalization.

\subsection{Case Study}
To intuitively investigate whether ReCo can discover the right problem when the causal chain is unreliable, we provide an example made by ReCo. As shown in Table~\ref{tab:case}, ``\emph{bacteria}'' in a cosmetic scene caused by ``\emph{acne}'' cannot lead to ``\emph{salmonellosis}''. ReCo gives the right label and the right problem which causes the causal chain unreliable. Refer to Appendix~\ref{app:case} for more cases.

\begin{table}[t]
\small
\centering
\begin{tabular}{lc}
\toprule
\multicolumn{2}{c}{production of sebum$\rightarrow$ acne $\rightarrow$ bacteria $\rightarrow$ salmonellosis} \\ \midrule
\textbf{ReCo Prediction}  & Unreliable     \\
\textbf{Scene Drift}      & True  \\
\textbf{Threshold Effect} & False \\ \bottomrule
\end{tabular}
\caption{An example made by ReCo. ReCo makes the right prediction and gives the reason why this chain breaks: \emph{``salmonellosis''} will not happen in the scene where \emph{``acne''} causes \emph{``bacteria''}.}
\label{tab:case}
\end{table}

\section{Related Work}
\subsection{Causal Knowledge Acquisition}
Causal knowledge is crucial for various artificial intelligence applications.
Many works~\cite{heindorf2020causenet,10.1145/3366423.3380107} extract large-scale and precise causal pairs through neural or symbolic ways. Hereafter, they connect causal pairs into causal chains or graphs based on the textual or semantic similarity between events~\cite{chang2004causal,ijcai2020-guided,hashimoto2014toward}. 

\citet{luo2016commonsense} used linguistic patterns~\cite{chang2004causal} to construct CausalNet. \citet{heindorf2020causenet} built CauseNet from web resources.
\citet{rashkin2018event2mind} constructed Event2mind and \citet{sap2019atomic} built Atomic both through crowd-sourcing. \citet{10.1145/3366423.3380107} proposed a large-scale eventuality knowledge graph called ASER. \citet{ijcai2020-guided} built CausalBank to improve the coverage of the causal knowledge base.

Previous studies mainly focused on extracting high-precision causal pairs, while ignoring the transitive problems when connecting event pairs into causal chains. We are trying to solve the two transitive problems in generating reliable causal chains.

\subsection{Causal Reasoning}
Causal reasoning aims at grasping the causal dependency between cause and effect, which consists of statistical-based and neural-based methods.

As for statistical-based methods, \citet{gordon2011commonsense} measured PMI based on a personal story corpus and then measured causality between words with PMI. \citet{luo2016commonsense} and \citet{sasaki2017handling} introduced direction information into causal strength index. Then they infer causality between events by combining the causality of word pairs. 

Many neural-based methods introduce the semantics of events to measure the causality of causal pairs. Of late, \citet{xie2019distributed} proposed to measure word-level causality with an attention-based mechanism. \citet{wang2019superglue} and \citet{li2019learning} finetuned the pre-trained language model to resolve causal reasoning task and achieve impressive results. \citet{li2021causalbert} injected a vast amount of causal pair knowledge into the pre-trained language model and got a noticeable improvement in COPA~\cite{roemmele2011choice} causal reasoning task. \citet{du2021excar} introduced evidence events to the causal pairs and used a conditional Markov neural logic network to model the causal paths between cause and effect events, to achieve stable and self-explainable causal reasoning. \citet{du-etal-2022-e} introduced general truth to event pair for investigating explainable causal reasoning.

Most of the above causal reasoning studies focus on causal pair reasoning, while we are trying to solve the reliable causal chain reasoning.

\section{Conclusion}

We explore the problem of causal chain reasoning and propose a novel framework called ReCo to overcome the two main transitive problems of threshold effect and scene drift.
ReCo first constructs an SCM for each causal chain, the SCM introduces exogenous variables to represent the threshold and scene factor of the causal pairs, and then conducts EA-CVAE to implicitly learn the representations of the exogenous variables with the contexts. Finally, ReCo devises SRNN to estimate the threshold and scene contradictions across the exogenous variables. Experiments show that ReCo can achieve the best CCR performances on both Chinese and English datasets. 

\section{Acknowledgments}
We would like to thank Bibo Cai and Minglei Li for their valuable feedback and advice, and the anonymous reviewers for their constructive comments, and gratefully acknowledge the support of the Technological Innovation “2030 Megaproject” - New Generation Artificial Intelligence of China (2018AAA0101901), and the National Natural Science Foundation of China~(62176079, 61976073).

\section{Limitations}
There may be some possible limitations in this study. First, the threshold and scene factors are hard to explicitly capture, which might hinder ReCo to achieve higher performances. Second, due to the loss function possessing three components and the nature of CVAE, it needs more attempts to reach convergence in training. Third, due to the nature of the CEG, each causal pair consists of only one context. Having multiple contexts for each causal pair would be better to cover more conditions as well as capture the threshold effect and scene drift problems more precisely. Moreover, it would be better to have larger CCR datasets. Future research should be undertaken to explore a more efficient and general model architecture as well as obtain larger Chinese and English CCR datasets with higher agreements and multiple contexts.

\bibliography{anthology}
\bibliographystyle{acl_natbib}

\appendix
\section{CEG Construction}
\label{app:ceg}
CEG (Chinese Event Graph)~\cite{ding2019elg} is a large-scale and open-domain causal event graph, which consists of more than 1.6 million events and 3.6 million cause-effect edges. We list the steps of constructing CEG as follows:

1) Crawling news documents from the news websites (such Netease news\footnote{https://news.163.com/}, Tencent news\footnote{https://news.qq.com/}, etc.).

2) Conducting causal pairs extraction through sequence labeling ($B_\text{Cause}, I_\text{Cause}, B_\text{Effect}, I_\text{Effect}, O$), training data are annotated through crowdsourcing.

3) Event similarity computation through Jaccard similarity coefficient~\cite{niwattanakul2013using} and event clustering using a threshold.

4) Extracting common elements from events (make sure that at least a verb and a noun are kept) in the same cluster to generalize events.

5) Connecting event pairs into causal chains and CEG.

\section{CCR Examples}
\label{sec:examples}
Examples of Chinese and English CCR are shown in Table~\ref{tab:chinese} and Table~\ref{tab:example}, respectively. $2$ and $4$ in the labels represent the causal chain that will meet problems at the second and the fourth causal relationship, respectively. And the problem types are threshold effect and scene drift for Chinese and English CCR examples, respectively.

\begin{table}[t]
\small
\centering
\begin{tabular}{l|l}
\toprule
\textbf{Events} &
  \begin{tabular}[c]{@{}l@{}} \textbf{A}: 销量下滑\\ \textbf{B}: 市场竞争加剧\\ \textbf{C}: 深圳发展\\ \textbf{D}: 城市化进程快\\ \textbf{E}: 水源水质差\end{tabular} \\ \midrule
\textbf{Contexts} &
  \begin{tabular}[c]{@{}l@{}}\textbf{A $\rightarrow$ B}: 销量下滑导致了终端市场竞\\争加剧\\ \textbf{B $\rightarrow$ C}: 通信市场竞争加剧将有助于\\深圳的通信设备业发展\\ \textbf{C $\rightarrow$ D}: 深圳的向西发展使得宝安的\\城市化进程越来越快\\ \textbf{D $\rightarrow$ E}: 水源水质极差的原因是周边\\城市化进程较快\end{tabular} \\ \midrule

\textbf{Label} &
  2 \\ \midrule
\textbf{Wrong Type} &
    Threshold Effect \\ \bottomrule
\end{tabular}
\caption{An example in the Chinese CCR dataset.}
\label{tab:chinese}
\end{table}

\begin{table}[t]
\small
\centering
\begin{tabular}{l|l}
\toprule
\textbf{Events}     & \begin{tabular}[c]{@{}l@{}}\textbf{A}: Tired at work\\ \textbf{B}: Relax\\ \textbf{C}: Playing games \\ \textbf{D}: Dispute\\ \textbf{E}: Sent off by a red card\end{tabular} \\ \midrule
\textbf{Contexts} &
  \begin{tabular}[c]{@{}l@{}}\textbf{A $\rightarrow$ B}: Tired at work makes me need\\ to relax at weekends.\\\textbf{B $\rightarrow$ C}: Tom wants to relax by playing\\ games.\\ \textbf{C $\rightarrow$ D}: Jack and Mike dispute because\\ of playing games.\\ \textbf{D $\rightarrow$ E}: David Silver gets a red card\\ because of the dispute with the referee.\end{tabular} \\ \midrule

\textbf{Label}      & 4                                                                                                                                  \\ \midrule
\textbf{Wrong Type} & Scene Drift                                                                                                                        \\ \bottomrule
\end{tabular}
\caption{An example in the English CCR dataset.}
\label{tab:example}
\end{table}

\section{Causal Knowledge Injection}
\label{app:cki}
We use the English CCR dataset for different knowledge injections: causal pair knowledge (BERT$_\text{P}$), unfiltered causal chain knowledge (BERT$_\text{C}$), and causal chain knowledge distilled by ReCo (BERT$_\text{R}$). All the models are based on BERT-base~\cite{devlin2019bert}.
\subsection{Knowledge Injection Settings}
For BERT$_\text{P}$, we split causal chains in English CCR datasets into causal pairs, and for each causal pair, we randomly sample a cause event or effect event from other causal pairs to obtain negative samples. The cause together with the effect event will be concatenated and sent into the pre-trained BERT, then we use the representation of \text{\emph{[CLS]}} token in the last hidden state for binary classification.

For BERT$_\text{C}$, we split causal chains in the English CCR dataset into causal chains of length 2 to 5. As for negative samples, for each causal chain, we sample an event from another causal chain to replace the first or last event of the causal chain. The events in a causal chain will be concatenated into a sequence and sent into the pre-trained BERT, then we use the representation of \text{\emph{[CLS]}} token in the last hidden state for binary classification.

For BERT$_\text{R}$, we split causal chains filtered by ReCo into causal chains of length 2 to 5. As for negative samples, for each causal chain, we sample an event from another causal chain to replace the first or last event of the causal chain. The events in a causal chain will be concatenated into a sequence and sent into the pre-trained BERT, then we use the representation of \text{\emph{[CLS]}} token in the last hidden state for binary classification.

\subsection{Knowledge Injection Details}
For all methods (BERT$_\text{P}$, BERT$_\text{C}$ and BERT$_\text{R}$), we use the base version of BERT~\cite{devlin2019bert}. The batch size is 36, and we use Adam~\cite{kingma2014adam} optimizer with the learning rate of 1$e$-5. All three models are pre-trained for 2 epochs.
\subsection{Downstream Tasks Finetuning}

\begin{table}[t]
\small
\centering
\begin{tabular}{llll}
\toprule
\textbf{Datasets}                  & \textbf{Train} & \textbf{Dev}  & \textbf{Test} \\ \midrule
Event StoryLine~v0.9 & 8,279  & 1,034 & 1,034 \\
BeCAUSE~2.1          & 1,741  & 216  & 216  \\
COPA                      & 450   & 50   & 500  \\
CommonsenseQA             & 9,741  & 1,221 & -  \\ \bottomrule
\end{tabular}
\caption{Statistics of Event StoryLine~v0.9~\protect\cite{caselli2017event}, BeCUASE~2.1~\protect\cite{dunietz2017because}, COPA~\protect\cite{roemmele2011choice}, CommonsenseQA~\protect\cite{talmor2019commonsenseqa} datasets.}
\label{tab:datasets}
\end{table}

\subsubsection{Dataset Settings}
\noindent$\bullet$ \textbf{Event StoryLine~v0.9}~\cite{caselli2017event} For the Event StoryLine~v0.9 dataset, we only keep the intra-sentence causal pairs and ensure that the cause event precedes the effect event. Finally, we randomly split the filtered causal pairs into train, dev, test sets.

\noindent$\bullet$ \textbf{BeCAUSE~2.1}~\cite{dunietz2017because} For the BeCAUSE~2.1 dataset, we first extract event pairs from the annotated data, then we manually split the event pairs into train, dev, test sets.

\noindent$\bullet$ \textbf{COPA}~\cite{roemmele2011choice} For the COPA dataset, for the reason that COPA does not have a training set, we randomly sample 90\% of the dev set for training, the remaining 10\% as the new dev set.

\noindent$\bullet$ \textbf{CommonsenseQA}~\cite{talmor2019commonsenseqa} For the CommonsenseQA dataset, we use the dev set for testing due to the test set of CommonsenseQA is a blind set.

The statistics of the four datasets are shown in Table~\ref{tab:datasets}.

\subsubsection{Finetuning}
We finetune BERT$_\text{O}$, BERT$_\text{P}$, BERT$_\text{C}$ and BERT$_\text{R}$ on the above four downstream tasks.

For Event StoryLine~v0.9 and BeCAUSE~2.1, we concatenate the event pair into a sequence and send it into the above four models, then the representation of \emph{[CLS]} in the last hidden state is used for binary classification. We use F1 score and accuracy as the evaluation metrics of Event StoryLine~v0.9 and BeCAUSE~2.1, respectively.

For COPA and CommonsenseQA tasks, we concatenate the premise (question) together with one of the hypotheses (alternatives) and feed it into all four models, then we use the \emph{[CLS]} token in the last hidden state for classification. We use accuracy as the evaluation metric for both COPA and CommonsenseQA.

As for the finetuning settings of the above four models, the batch size is set to 40, and we use Adam~\cite{kingma2014adam} optimizer with the learning rate of 1$e$-5. An early-stopping mechanism is applied for finetuning.

\section{Ablation Study}
\label{ap:ab}
\subsection{EA-CVAE}
For investigating the importance of EV-CVAE in ReCO, we remove the EA-CVAE component in ReCo, and for constructing the SCM~\cite{pearl2009causality}, we use the contexts of the causal pairs as the exogenous variables in the SCM. Other components of ReCo are not changed and the training settings are the same as the original ReCo.

\subsection{Problems Estimators}
We devise two problem estimators to estimate threshold effect and scene drift problems. For investigating the importance of these two problem estimators, we remove the supervised signal (by removing $L_\text{logic}$ in the loss) of Threshold and Scene Estimators in the SRNN, the parameters of the two problem estimators are only tuned by the final reliability prediction task (note that EA-CVAE are kept for training, and the tuning of Kullback-Leibler divergence loss~\cite{hershey2007approximating} will not change the parameters in the two problem mechanisms). The model architecture and the training settings of this setting are the same as the original ReCo.

\subsection{Logic Loss}
The logic loss is used to apply a logic constraint on the predictions of ReCo. When the causal chain is reliable, both the threshold effect and scene drift problems do not exist. Moreover, when the causal chain is unreliable, one of the transitive problem should exist. For investigating the effect of the logic loss, we replace the logic loss with two cross entropy losses. One of the cross-entropy loss is conducted to supervise the threshold effect problem, and the other is used to supervise the scene drift problem.

\section{Case Study}
\label{app:case}
\begin{table}[!t]
\small
\centering
\begin{tabular}{lc}
\toprule
\multicolumn{2}{c}{reading$\rightarrow$ myopia $\rightarrow$ problems $\rightarrow$ stress} \\ \midrule
\textbf{ReCo Prediction}  & Unreliable     \\
\textbf{Scene Drift}      & False  \\
\textbf{Threshold Effect} & True \\ \bottomrule
\end{tabular}
\caption{An example made by ReCo. Reco makes the right prediction and gives the reason why this chain is unreliable. ``\emph{Problems}'' conditioned on ``\emph{reading}'' and ``\emph{myopis}'' is not enough to lead to ``\emph{stress}''.}
\label{tab:case1}
\end{table}

\begin{table}[t]
\small
\centering
\resizebox{\linewidth}{!}{
\begin{tabular}{lc}
\toprule
\multicolumn{2}{c}{volume growth$\rightarrow$ revenue growth $\rightarrow$ improvement $\rightarrow$ energy savings} \\ \midrule
\textbf{ReCo Prediction}  & Unreliable     \\
\textbf{Scene Drift}      & True  \\
\textbf{Threshold Effect} & False \\ \bottomrule
\end{tabular}}
\caption{An example made by ReCo. Reco makes the right prediction and gives the reason why this chain is unreliable. ``\emph{Energy savings}'' will not happen in the scene of ``\emph{volume growth}''$\rightarrow$``\emph{revenue growth}'' $\rightarrow$``\emph{improvements}''.}
\label{tab:case2}
\vspace{-0.3cm}
\end{table}

We provide another two English examples predicted by ReCo. The examples of threshold effect and scene drift problems are shown in Table~\ref{tab:case1} and Table~\ref{tab:case2}, respectively.

\end{CJK}
\end{document}